\pdfoutput=1

\documentclass[11pt]{article}

\usepackage{ACL2023}

\usepackage{times}
\usepackage{latexsym}

\usepackage[T1]{fontenc}

\usepackage[utf8]{inputenc}

\usepackage{microtype}

\usepackage{inconsolata}
\usepackage[normalem]{ulem}
\usepackage{soul}
\usepackage{multicol}
\usepackage{multirow}
\usepackage{pgfplots}
\usepackage{graphicx}
\usepackage{ctable}
\usepackage{subfig}

\usepackage{amsthm,amsmath,amsfonts,bm,xspace}
\usepackage{upgreek}
\usepackage{color}

\def\vs{{\em v.s.}\xspace}

\def\eqref#1{(\ref{#1})}

\def\1{\bm{1}}

\def\vd{{\bm{d}}}

\def\vh{{\bm{h}}}

\def\vs{{\bm{s}}}

\def\vz{{\bm{z}}}

\DeclareMathAlphabet{\mathsfit}{\encodingdefault}{\sfdefault}{m}{sl}
\SetMathAlphabet{\mathsfit}{bold}{\encodingdefault}{\sfdefault}{bx}{n}

\newcommand{\E}{\mathbb{E}}

\title{\textsc{NormMark}: A Weakly Supervised Markov Model for \\ 
Socio-cultural Norm Discovery}

\author{Farhad Moghimifar \and Shilin Qu \and Tongtong Wu \\
          \textbf{Yuan-Fang Li} \and \textbf{Gholamreza Haffari} \\
         Department of Data Science and AI, Monash University, Australia \\
         \texttt{\{first.lastname\}@monash.edu}}

\begin{document}
\maketitle
\begin{abstract}
Norms, which are culturally accepted guidelines for behaviours, can be integrated into conversational models to generate utterances that are appropriate for the socio-cultural context. 
Existing methods for norm recognition tend to focus only on surface-level features of dialogues and do not take into account the interactions within a conversation. 
To address this issue, we propose \textsc{NormMark}, 
a probabilistic generative Markov model 
to carry the latent features throughout a dialogue.
These features are captured by discrete and continuous latent variables   conditioned on the conversation history, and improve the model's ability in  norm recognition. 
The model is trainable on weakly annotated data using the variational technique. 
On a dataset with limited  norm annotations, we show that our approach achieves higher F1 score,  outperforming current state-of-the-art methods, including  GPT3. 


\end{abstract}

\section{Introduction}

Norms can be thought of as pre-defined socio-culturally acceptable boundaries for human behaviour~\citep{fehr2004social}, and incorporating them into conversational models helps to produce contextually, socially and culturally appropriate utterances. For instance, identifying the socio-cultural norm of \emph{Greeting} in a negotiation helps to generate responses suitable for the power dynamics and social setting. Whereas, failing to detect and adhere to such norms can negatively impact social interactions~\citep{hovy-yang-2021-importance}. Recent advances in developing chatbots have also highlighted the necessity of incorporating such implicit socio-cultural information into machine-generated responses, in order to approximate human-like interactions~\citep{huang2020challenges, liu-etal-2021-towards}.

\begin{figure}[t]
    \centering
    \includegraphics[width = 0.8\linewidth]{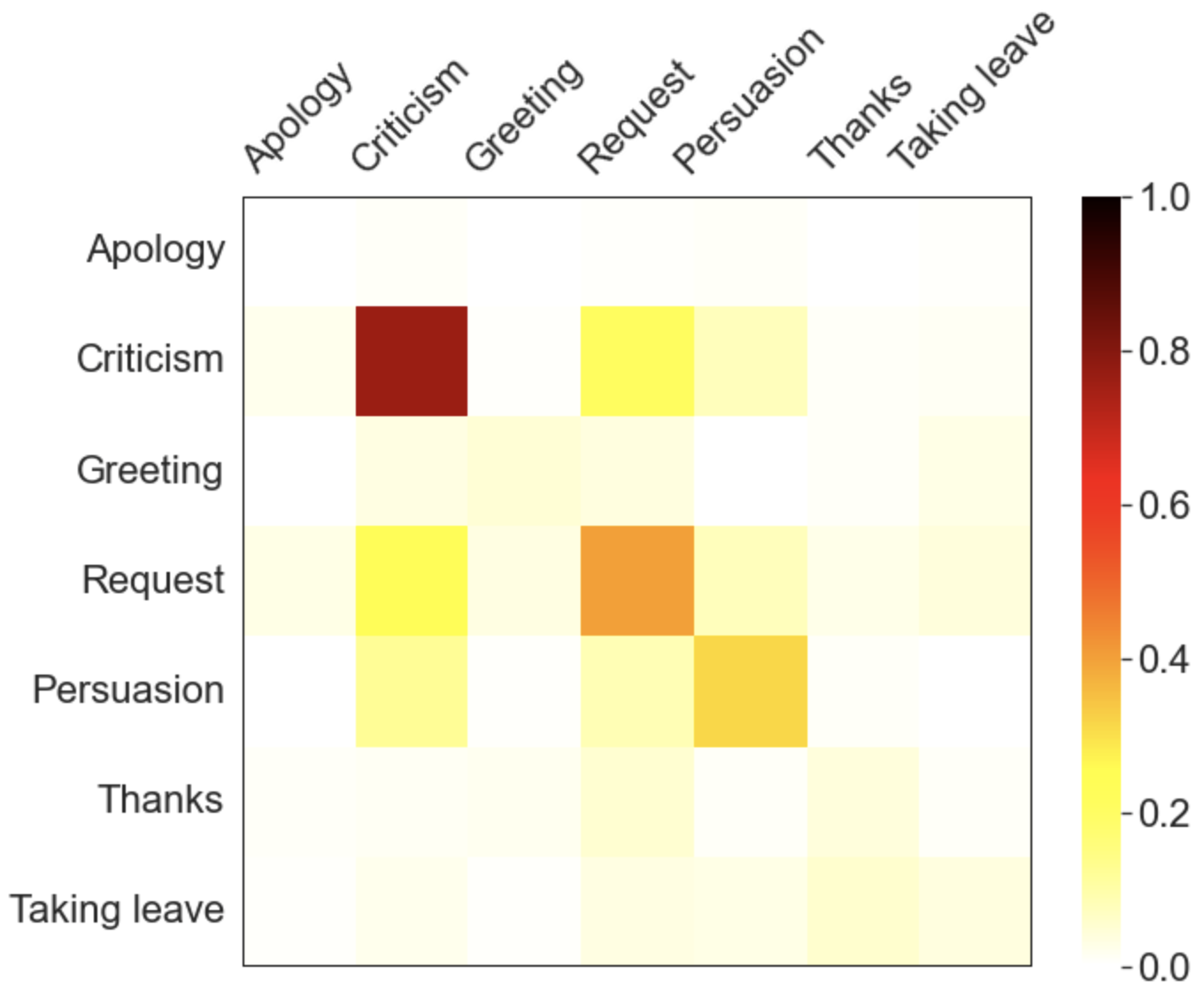}
    \caption{The heatmap of norm distribution based on norm label of the previous segment, constructed from LDC2022E20. Unit in column $j$ of row $i$ shows the probability of norm $i$ following norm $j$.}
    \label{fig:norm-dist}
\end{figure}

Norm discovery is a nascent research problem, and current approaches~\citep{hwang2021comet} heavily rely on manually constructed sets of rules from available resources such as Reddit~\citep{forbes-etal-2020-social,ziems-etal-2022-moral}. 
In addition to the time and cost inefficiency of such approaches, the construction and use of these banks of norms treat each sentence or segment in isolation, and they fail to take the dependencies between norms in the flow of a dialogue into account~\citep{fung2022normsage,chen2021weakly}.
For instance, it is most likely that a dialogue segment containing the norm of \emph{Request} follows a segment that includes ~\emph{Request} and \emph{Criticism}~(Figure~\ref{fig:norm-dist}). Furthermore, such approaches require a large amount of annotated data,  limiting their performance on sparsely labelled resources.

To address these limitations, in this paper, we propose a deep generative Markov model that captures the inter-dependencies between turns (segments) of partially-labelled dialogues. 
The model includes  two types of latent variables (LVs): (i) the discrete LVs  capture the socio-cultural norms of the dialogue turns, and (ii) the continuous LVs capture other aspects, e.g. related to fluency, topic, and meaning. 
These latent variables facilitate capturing label- and content-related properties of the previous turns of the conversation, and are conditioned on the previous turns in a Markovian manner.  
We train the model on weakly annotated data using the variational technique, building on variational autoencoders  \cite{Kingma2014}. 

To evaluate the performance of our model in the task of socio-cultural norm discovery, we conducted experiments on an existing dataset. Experimental results show superiority of our model, by 4 points in F1 score, over the state-of-the-art approaches, in which each segment of a dialogue is modelled independently of the others. Furthermore, by evaluating our model on low amounts of training data, we show the capability of our proposed approach in capturing socio-cultural norms on partially-labeled data.

\begin{figure}[t]
	   \centering
	   \includegraphics[width=0.7\linewidth]{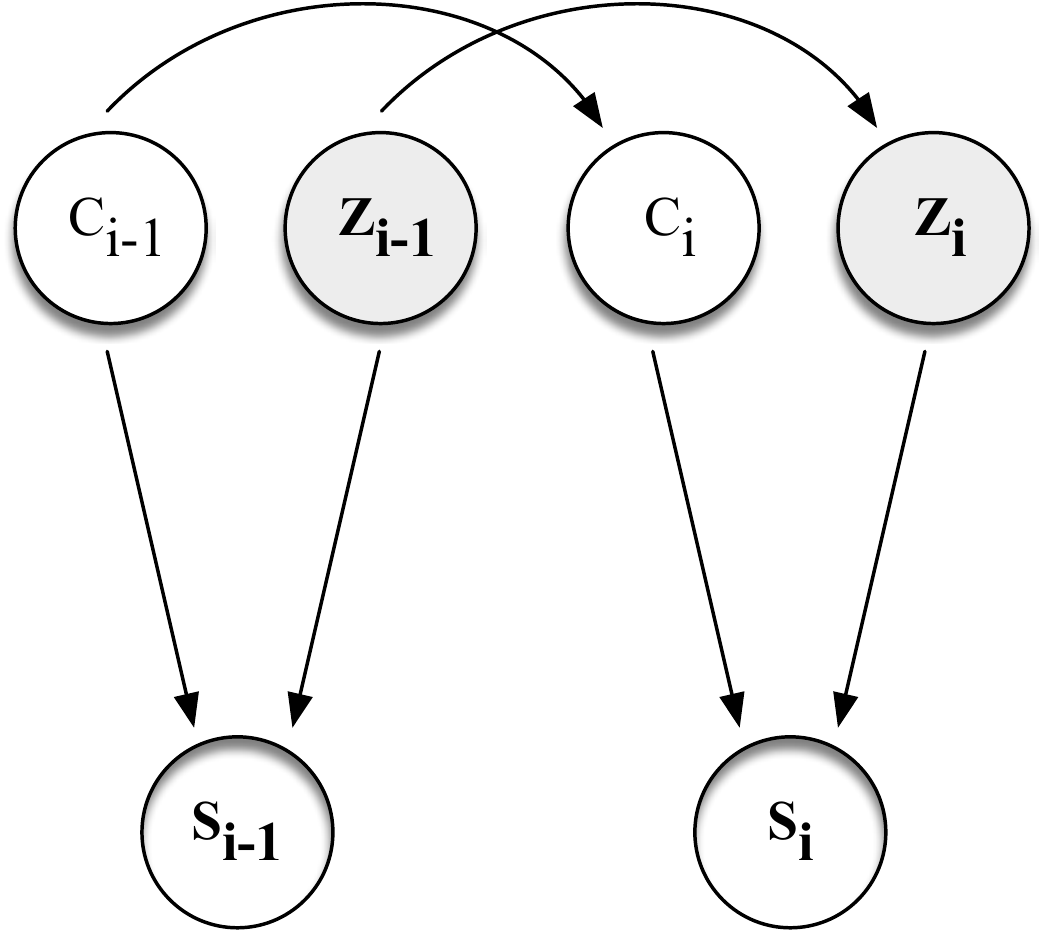}
\caption{A graphical representation of our probabilistic generative model $\textsc{NormMark}$.
}
\label{fig:g-model}
\end{figure}

\section{Related Works}
Recent approaches have tried to develop models with the human psychological and behavioural capabilities~\citep{jiang2021delphi,botzer2022analysis,lourie2021scruples}. Other approaches targeted identifying implicit social paradigms by developing sequence generation models~\citep{moghimifar-etal-2020-cosmo,bosselut-etal-2019-comet}. However, the task of socio-cultural norm discovery has been overlooked, mostly due to the lack of proper annotated data~\citep{fung2022normsage}. \citet{forbes-etal-2020-social} present a dataset of social norms, collected from Reddit and propose a generative model to expand this collection. 
\citet{zhan2022let} also showed how social norms can be useful in conducting better negotiation dialogues.
In a similar approach, \citet{ziems-etal-2022-moral} present a corpus of moral norms. \citet{zhan2023socialdial} and  \citet{fung2022normsage} use a prompt-based large-scale language model to generate rules from dialogues. More similar to our approach, existing models identify labels associated with utterances of dialogues~\cite{chen2021weakly,yang-etal-2019-lets, yu-etal-2020-ch}. However, these approaches fail to take into account the flow of contextual information throughout a dialogue. In contrast to these studies, our approach addresses this task by considering the inter-dependencies between turns of dialogues.

\section{A Generative Markov Model for Socio-Cultural Norm Discovery}

We are given a set of dialogues $\mathcal{D} = \{\vd^i\}_{i=1}^n$, where each dialogue consists of a set of turns (or segments) $\vd^i = \{ \vs_j^i\}_{j=1}^m$. Each turn consists of a sequence of tokens from a vocabulary set $\mathcal{V}$. The dialogue set $\mathcal{D}$ consists of two subsets of labeled~($\mathcal{D}_L$) and unlabeled~($\mathcal{D}_U$) dialogues, where each turn $\vs_m^i \in \mathcal{D}_L$ is annotated with a socio-cultural norm label $c_i \in \mathcal{C}$ with a total of $K$ norm classes. The turns in the unlabeled dataset lack socio-cultural norm labels. Our goal is to develop a model that, by using contextual information carried from previous turns of the dialogue, discovers the socio-cultural norm associated with the turns of a dialogue.

\paragraph{Probabilistic Generative Model.}
Our model (shown in Fig.~\ref{fig:g-model}) assumes a directed generative model, in which a turn is generated by a factor capturing the socio-cultural norms and another factor capturing other aspects, e.g. topic and syntax.
For each turn, the socio-cultural norm factor is captured by a discrete latent variable $c_i$, and the other aspects are captured by a  continuous latent variable $\vz_i$. 
As our aim is to leverage the contextual information, the latent variables of each turn of the dialogue are conditioned on those from the previous turn in Markovian manner. 
As such, our proposed generative model for each turn is as follows:

\begin{align*}
    &p_{\theta}(\vs_i,\vz_i,c_i|\vz_{i-1},c_{i-1}) = \\
    &p_{\theta}(\vs_i|c_i,\vz_i) p_{\theta}(\vz_i|\vz_{i-1}) p_{\theta}(c_i|c_{i-1})
\end{align*}
where 
$p_{\theta}(c_i|c_{i-1})$ and $p_{\theta}(\vz_i|\vz_{i-1})$ capture the dependency of the causal factors on the previous turn, and $p_{\theta}(\vs_i|c_i,\vz_i)$ is a sequence generation model conditioned on the causal factors. 

\paragraph{Training.} To train the model, the likelihood function for a  dialogue in $D_U$ is:
\begin{align*}
    & p_{\theta}(\vs_{1}..\vs_n) = \sum_{c_{1}..c_n} \int d(\vz_{1})..d(\vz_n) \times \\
    & \prod_{i=1}^n p_{\theta}(\vs_i|c_i,\vz_i) p_{\theta}(\vz_i|\vz_{i-1}) p_{\theta}(c_i|c_{i-1}).
\end{align*}
%
Intuitively, the training objective for each dialogue turn corresponds an extension of the variational autoencoder (VAE) which involves: (i)  both discrete and continuous latent variables, and (ii) conditioning on the latent variables of the previous turn. As such, we resort to the following variational evidence lowerbound (ELBO) for the unlabeled turns:  
\begin{align*}
& \log p(\vs_i|c_{i-1},\vz_{i-1}) \ge \E_{q_{\phi} (c_i|\vs_i,c_{i-1})} \big\{\\
& \E_{q_{\phi}(\vz_i|\vs_i,\vz_{i-1})} \big[\log p_{\theta}(\vs_i|\vz_i,c_i)\big] \big\}  \\
&- KL[q_{\phi}(\vz_i|\vs_i,\vz_{i-1})||p_{\theta}(\vz_i|\vz_{i-1})]  \\
& - KL[q_{\phi}(c_i|\vs_i,c_{i-1})||p_{\theta}(c_i|c_{i-1})]
\end{align*}
where $q_{\phi}$'s are variational distributions. We have nested ELBOs, each of which corresponds to a turn in the dialogue. We refer to the collection of these ELBOs for all dialogues in $D_U$ by $\mathcal{L}(D_U)$.
For the labeled turns, the ELBO for a dialogue turn is,
\begin{align*}
& \log p(\vs_i,c_i|c_{i-1},\vz_{i-1}) \ge \log p_{\theta}(c_i|c_{i-1})  \\
&+ \E_{q_{\phi}(\vz_i|\vs_i,\vz_{i-1})} \big[\log p_{\theta}(\vs_i|\vz_i,c_i)\big]   \\
&- KL[q_{\phi}(\vz_i|\vs_i,\vz_{i-1})||p_{\theta}(\vz_i|\vz_{i-1})]  
\end{align*}
where we also add the term $\log q_{\phi}(c_i|\vs_i,c_{i-1})$ to the training objective. We refer to the collection of ELBOs for all dialogues in the labeled data as $\mathcal{L}(D_L)$. Finally, the training objective based on the labeled and unlabeled dialogues is $\mathcal{L}=\mathcal{D_U} + \lambda \mathcal{D_L}$, where $\lambda$ trades off the effect of the labeled and unlabeled data. We resort to the reparametrisation trick for continuous and discrete (Gumble-softmax~\citep{jang2017categorical}) latent variables when optimising the  training objective. 

\paragraph{Architectures.} 
We use 
a transformer-based encoder to encode the turns $\vs_i$ with a hidden representation $\vh^s_i$. 
The \emph{classifier} $q_{\phi}(c_i|\vs_i,c_{i-1})$ is a 2-layer MLP with \texttt{tanh} non-linearity whose inputs are $\vh^s_i$ and the embedding of $c_{i-1}$.
For $q_{\phi}(\vz_i|\vs_i,\vz_{i-1})$, we use a a multivariate Gaussian distribution, whose parameters are produced by MLPs from $\vh^s_i$ and $\vz_{i-1}$. 
For $p_{\theta}(\vs_i|\vz_i,c_i)$, we use an LSTM decoder, where this is performed by replacing pre-defined special tokens in the embedding space with $\vz_i$ and $c_i$. 
For $p_{\theta}(c_t|c_{t-1})$, we use MLP with a softmax on top.

%

\section{Experiments}
In this section we report the performance of our model on the task of socio-cultural norm discovery in comparison to the current state-of-the-art models.

\paragraph{Dataset} In our experiments, we use LDC2022E20. This dataset consists of 13,074 segments of dialogues in Mandarin Chinese. The dialogues are from text, audio, and video documents, where we transcribed the audio and video files using Whisper~\citep{radford2022robust}. The segments have been labelled from the set of socio-cultural norm labels of \emph{none}, \emph{Apology}, \emph{Criticism}, \emph{Greeting}, \emph{Request}, \emph{Persuasion}, \emph{Thanks}, and \emph{Taking leave}. We split the data into train/test/development sets with the ratio of 60:20:20. Each dialogue is divided into sequences of segments of length 5, where on average each segment consists of 8 sentences.
We report the performance of our model, in comparison to the baselines, when using the maximum number of labeled data in the training set~(Max).
In addition, to evaluate the effect of the amount of training data on the performance of our model, we randomly select 50 and 100 of these sequences of dialogues for training, and report the results on the test set.

\paragraph{Baselines.} We compare our model with LSTM~\citep{hochreiter1997long} and BERT~\citep{devlin-etal-2019-bert}, where each turn of a dialogue is encoded separately. We use WS-VAE-BERT~\citep{chen2021weakly} as another baseline, which encodes the contextual representation of a segment via a latent variable. However, WS-VAE-BERT does not capture the connections between segments. To experiment with the performance of our model on limited labeled data, we compare it to SetFit~\citep{tunstall2022efficient}, which has proven to be a strong few-shot learning model. 
Similar to our model, we use \texttt{`bert-base-chinese'} as the backbone of BERT and WS-VAE-BERT, and \texttt{`sbert-base-chinese-nli'} has been used in SetFit.
Additionally, we compare our model with a prompt-base large-scale language model GPT-3 \texttt{text-davinci-003}~\citep{brown2020language} and ChatGLM~\citep{du2022glm}, where the norm labels are given to the model with segments of dialogue, and the model is asked to generate a socio-cultural norm label from the list.

\paragraph{Evaluation Metrics.} Following previous works in classification tasks, we report the macro averaged precision, recall, and F1 score of the models in predicting the socio-cultural norm label of each segment of a dialogue.

\subsection{Results}
Table~\ref{tab:main-results} summarises the main results of the conducted experiment on LDC2022E20 data. On Max setting, where the model uses the maximum number of datapoints in the training set, our model outperforms all of the baselines with a margin of 4 and 6 points on F1 and precision, respectively, and achieves a comparable result in recall. This gap between our model and WS-VAE-BERT indicates the effect of carrying contextual information from previous turns of conversation. In addition, lower results of GPT-3 suggest that discovering socio-cultural norms is a challenging task, which needs higher-level reasoning. 

\begin{table}[t]
    \centering
    \resizebox{\linewidth}{!}{
    \begin{tabular}{l   c  c  c  c}
             &  \multicolumn{3}{c}{\textbf{Max}} & \\
            \toprule[0.1em]
            \textbf{Model} &  \textbf{P} & \textbf{R} & \textbf{F1}  & Size\\
            \toprule[0.1em]
            LSTM &   9.13 & 12.57 & 10.08 & 0.4M\\
            BERT &   38.42 & 32.33 & 33.34 & 109M\\
            WS-VAE-BERT &   42.03 & \textbf{40.74} & 39.01 & 132M\\
            SetFit &  41.42 & 40.23 & 40.54 & 102M\\
            \hline
            ChatGLM & 17.64 & 20.55 & 17.19 & 6B \\
            GPT-3 &   39.86 & 35.05 & 33.61 & 175B\\
            \hline
            $\textsc{NormMark}_{\texttt{zero}}$  & 44.14 & 36.97 & 39.49 & 136M\\
            \textsc{NormMark} &  \textbf{47.92} & 38.67 & \textbf{44.20} & 131M\\
            \toprule[0.1em]
    \end{tabular}}
    \caption{Segment-level socio-cultural norm prediction performance~(precision, recall and F1 score). The results are reported by training the models on maximum number of labelled sequences of dialogues.}
    \label{tab:main-results}
\end{table}

\begin{table}[b]
    \centering
    \resizebox{0.9\linewidth}{!}{
    \begin{tabular}{l c c c}
         \textbf{Model} &  \textbf{50} &  \textbf{100} & \textbf{Max} \\
         \hline
       $\textsc{NormMark}_{\texttt{zero-extended}}$ & 13.41 & 14.33 & 20.33 \\
       $\textsc{NormMark}_{\texttt{extended}}$ & 13.48 & 14.76 & 20.25 \\
       \hline
       \textsc{NormMark} & 32.46 & 34.43 & 44.2 \\
       \hline
    \end{tabular}}
    \vspace{-1ex}
    \caption{Segment-level socio-cultural norm prediction of two variation of our  approach, in comparison to our model. The results are macro-averaged F1 score.}
    \label{tab:extended}
\end{table}

\paragraph{Amount of Labeled Data.}
To evaluate the performance of our model with less amount of training data, we report the results on using only 50 and 100 datapoints during training, in Table~\ref{tab:data-size}. When using 100 sequences of turns, our model achieves the highest score in F1, and improves the precision and recall by more than 3 points over non-prompt based models. However, GPT-3 outperforms our proposed model in these two metrics. Similarly, on a more limited number of training data~(setting 50), GPT-3 shows its dominance. Nevertheless, our model performs the best amongst the other baselines, by improving the F1 score by 3 points.
\begin{table}[ht]
    \centering
    \resizebox{\linewidth}{!}{
    \begin{tabular}{l    c  c  c  c  c  c }
            & \multicolumn{3}{c}{\textbf{50}} & \multicolumn{3}{c}{\textbf{100}} \\
            \toprule[0.1em]
            \textbf{Model} &  \textbf{P} & \textbf{R} & \textbf{F1} & \textbf{P} & \textbf{R} & \textbf{F1} \\
            \toprule[0.1em]
            LSTM &  7.92 & 12.5 & 9.69 & 11.06 & 12.58 & 9.90 \\
            BERT &  10.85 & 15.74 & 12.58 & 10.46 & 15.50 & 11.82  \\
            WS-VAE-BERT &  20.43 & 17.21 & 16.60 & 36.38 & 22.48 & 23.37 \\
            SetFit & 30.42 & 26.25 & 27.86 & 32.12 & 30.54 & 31.32\\
            \hline
            ChatGLM & 17.64 & 20.55 & 17.19 & 17.64 & 20.55 & 17.19 \\
            GPT-3 &  \textbf{39.86} & \textbf{35.05} & \textbf{33.61} & \textbf{39.86} & \textbf{35.05} & 33.61 \\
            \hline
            $\textsc{NormMark}_{\texttt{zero}}$ &  25.94 & 19.71 & 20.04 & 35.41 & 27.73 & 28.33 \\
            \textsc{NormMark} &  \textbf{32.46} & \textbf{30.02} & \textbf{30.72} & \textbf{36.41} & \textbf{33.48} & \textbf{34.43} \\
            \toprule[0.1em]
    \end{tabular}}
    \caption{Segment-level socio-cultural norm prediction performance~(precision, recall and F1 score). The results are reported by training the models on 50, 100 number of labelled sequences of dialogues.}
    \label{tab:data-size}
\end{table}

\paragraph{Conditioning on the Context.} To analyse the effect of carrying contextual information from previous turns of dialogue, we report the performance of the simplified version of our model~($\textsc{NormMark}_{\texttt{zero}}$), where the connections from previous turn are omitted. As can be seen in Table~\ref{tab:main-results}, in all of the settings \textsc{NormMark} outperforms the simplified version, indicating the importance of inter-dependencies between turns. Furthermore, we developed two variations of $\textsc{NormMark}$ and $\textsc{NormMark}_{\texttt{zero}}$ where the contextual information from previous turns is carried directly through the previous segment~(Figure~\ref{fig:variations}). In Table~\ref{tab:extended}, the lower performance of these models suggests that the contextual information from the previous turn overshadows the representation of the latent variable as well as the norm label, and consequently the norm classifier is profoundly biased towards the previous turn of dialogue.

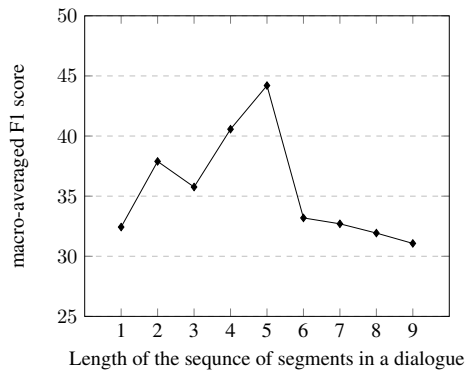
\begin{figure}[b]
    \centering
    \begin{tikzpicture}[scale = 0.7]
        \begin{axis}[
        xlabel={Length of the sequnce of segments in a dialogue},
        ylabel={macro-averaged F1 score},
        xmin=0, xmax=10,
        ymin=25, ymax=50,
        xtick = data,
        xticklabels={1,2,3,4,5,6,7,8,9},
        ytick={},
        ymajorgrids=true,
        grid style=dashed,
        ]
        \addplot 
        [color=black, mark=diamond*, mark options={scale=1}]
	coordinates {(1,32.43) (2,37.89)
		           (3,35.76) (4,40.57)
                      (5,44.2) (6,33.19)
                      (7,32.7) (8,31.93)
                      (9,31.07) };
        \end{axis}
        \end{tikzpicture}
    \caption{The performance of \textsc{NormMark} with different length of sequence of segments.}
    \label{fig:seq-len}
\end{figure}
\paragraph{Markov Order.} We further analysed the effect of carrying contextual meaning from previous turns of dialogues, by varying the size of the Markov conditioning context $l$  from 1 to 9, i.e. each of our proposed latent variables is conditioned on previous $l$ turns of dialogue. 

Figure~\ref{fig:seq-len} summarises the results. It shows that shorter context results in lower performance, due to passing less contextual information to the next turns. On the other hand, too long context results in lower performance as well, due to extra complexity of modelling longer dependencies in latent variables and norm labels. As shown in the figure, our model performs best with a context size of 5 on this dataset. 


\section{Conclusion}
In this work, we address the task of socio-cultural norm discovery from open-domain conversations. We present a probabilistic generative model that captures the contextual information from previous turns of dialogues. Through empirical results, we show that our model outperforms state-of-the-art models in addressing this task.

\section{Limitations}
We have studied the task of socio-cultural norm discovery based LDC2022E20 dataset, which consists of everyday situational interactions in Mandarin Chinese. Although we believe that our approach can used in other cultural settings, the current state of the model might not be generalisable to other cultures, unless further tuning is possible. 
Our model's ability in discovering such norms can help to improve conversational agents, however, real-world scenarios involving duplicitous or ambiguous terms might confuse our proposed approach. In addition, our model is limited to the textual modality, and we believe incorporating audio and visual features into the model can improve identifying socio-cultural norms.
Nonetheless, the reliance of our model on large-scale pre-trained language models might result in some deployment challenges in situations with limited resources.
Besides, all the reported results are by fixing a random seed running all experiments once.
\section{Ethics Statement}
Our work leverages pre-trained language models~(BERT), therefore similar potential risks of this model is inherited by our work.
\section{Acknowledgements}
This material is based on research sponsored by DARPA under agreement number HR001122C0029. The U.S. Government is authorised to reproduce and distribute reprints for Governmental purposes notwithstanding any copyright notation thereon. The authors are grateful to the anonymous reviewers for their helpful comments.

\vspace{-3ex}
\bibliography{anthology,custom}
\bibliographystyle{acl_natbib}

\appendix
\begin{figure*}[ht]
    \centering
    \includegraphics[width=\textwidth]{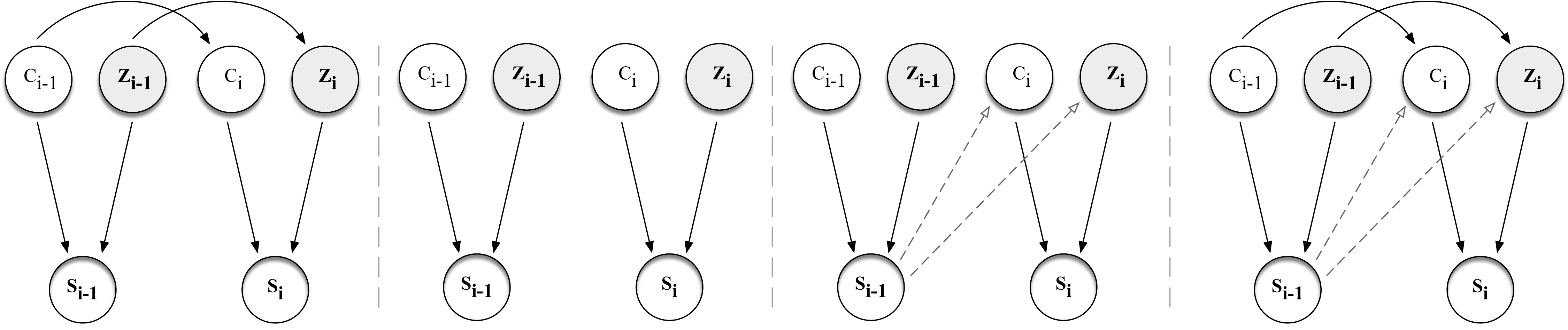}
    \vspace{-2ex}
    \caption{A graphical representation of our probabilistic generative model $\textsc{NormMark}$. The second model from left is a simplified version of our proposed approach where the contextual information from previous turns of dialogue is not carried through the current step~($\textsc{NormMark}_\texttt{zero}$). The next two models are \texttt{extended} versions of $\textsc{NormMark}_\texttt{zero}$ and $\textsc{NormMark}$, respectively, where direct contextual information from previous segment is carried to the current turn.}
    \vspace{-2ex}
    \label{fig:variations}
\end{figure*}
\vspace{-3ex}
\section{Experimental Details}
To train our model, we have used the pre-trained \texttt{`bert-base-chinese'}, which is licensed free to use for research purposes, as the encoder~\citep{kenton2019bert}, and we have used LSTM~\citep{hochreiter1997long} with hidden dimension of 128 and one hidden layer as the decoder. We used a dropout of 0.6 over the input. We implemented the norm classifier with a 2-layers MLP with \texttt{tanh} non-linearity on top. We used CrossEntropyLoss~\citep{zhang2018generalized} as loss function over the predictions of our model. We used AdamW~\citep{loshchilov2018decoupled} as the optimiser with the learning rate of 1e-5 for the encoder and 1e-3 for the rest of the network. We trained our model for 50 epochs, on a single machine with NVIDIA one A100 gpu, with an early stop if the validation accuracy is not improved for more than 20 iterations.

For the baselines, we have developed a network with a two-stacked LSTM layers followed by two linear layers. We compared out model with BERT, where uses the \texttt{`bert-base-chinese'} pre-trained model. Each of these two models where trained for 100 epochs, using AdmaW optimiser with the learning rates of 1e-3 and 5e-5, respectively. For WS-VAE-BERT~\citep{chen2021weakly}, we followed the source code provided in the paper. For replicating the document level labels, when a segment within the sequence of segments contained a socio-cultural norm, we labeled them 1, otherwise 0. We trained SetFit~\citep{hong2022tess} by following the online instructions on their GitHub repository~\footnote{https://github.com/huggingface/setfit}. Figure~\ref{fig:variations} shows the variations of our model, which we used in for the ablation study.

GPT-3~\citep{brown2020language} was used by incorporating the list of socio-cultural norms into the prompt as well as dialogues, and asking to generate the corresponding label. Our experiments on GPT-3 showed that using random examplars from the training set of LDC2022E20 results in a decrease in the performance. The LDC2022E20 dataset is the copyrighted property of (c) 2022 Trustees of the University of Pennsylvania and has been used for research purposes in CCU program. This dataset was developed to help models to identify socio-cultural norms in courses of dialogues.

In all of our experiments we used a fix random seed, hence all results are reported based on single-run of the models.

\end{document}